%% file: aaai23.tex
\documentclass[letterpaper]{article} 
\usepackage{aaai23}  
\usepackage{times}  
\usepackage{helvet}  
\usepackage{courier}  
\usepackage[hyphens]{url}  
\usepackage{graphicx} 
\usepackage{natbib}  
\usepackage{caption} 
\usepackage[title]{appendix}
\usepackage{algorithm}
\usepackage{algorithmic}

\usepackage{booktabs}
\usepackage{multirow}
\usepackage{caption}
\usepackage{graphicx} %
\usepackage{subfig}
\usepackage{amsmath}
\usepackage{newfloat}
\usepackage{listings}
\DeclareCaptionStyle{ruled}{labelfont=normalfont,labelsep=colon,strut=off} 
\floatstyle{ruled}
\newfloat{listing}{tb}{lst}{}
\floatname{listing}{Listing}
\title{\textsc{M-Sense}: Modeling Narrative Structure in Short Personal Narratives Using Protagonist's Mental Representations}
\author{
    Prashanth Vijayaraghavan, 
    Deb Roy
}
\affiliations{
    \textsuperscript{\rm 1} MIT Media Lab\\

    75 Amherst Street,\\
    Cambridge, MA, 02139 USA\\
    pralav@mit.edu, dkroy@media.mit.edu
}

\usepackage{bibentry}

\begin{document}

\maketitle
\begin{abstract}
Narrative is a ubiquitous component of human communication. Understanding its structure plays a critical role in a wide variety of applications, ranging from simple comparative analyses to enhanced narrative retrieval, comprehension, or reasoning capabilities. Prior research in narratology has highlighted the importance of studying the links between cognitive and linguistic aspects of narratives for effective comprehension. This interdependence is related to the textual semantics and mental language in narratives, referring to characters’ motivations, feelings or emotions, and beliefs. However, this interdependence is hardly explored for modeling narratives. In this work, we propose the task of automatically detecting prominent elements of the narrative structure by analyzing the role of characters’ inferred mental state along with linguistic information at the syntactic and semantic levels. We introduce a \textsc{Stories} dataset of short personal narratives containing manual annotations of key elements of narrative structure, specifically climax and resolution. To this end, we implement a computational model that leverages the protagonist’s mental state information obtained from a pre-trained model trained on social commonsense knowledge and integrates their representations with contextual semantic embed-dings using a multi-feature fusion approach. Evaluating against prior zero-shot and supervised baselines, we find that our model is able to achieve significant improvements in the task of identifying climax and resolution.
\end{abstract}

\section{Introduction}

Narratives are the fundamental means by which people organize, understand, and explain their experiences in the world around them. Researchers in the field of psychology maintain that the default mode of human cognition is a narrative mode \cite{beck2015life}. Humans share their personal experiences by picking specific events or facts and weaving them together to make meaning. These are referred to as personal narratives, a form of autobiographical storytelling that gives shape to experiences. \citet{polkinghorne1988narrative} suggested that personal narratives, like other stories, follow broad characteristics involving: (a) typically a beginning, middle, and end, (b) specific plots with different characters and settings, or events. Often, characters learn something or change as a result of the situation or a conflict and resolution, but not always. Some of these characteristics provide the basis for the organizational framework of a story, commonly referred to as the narrative structure or the storyline. The growing amount of personal narrative text information in the form of social media posts, comments, life stories, or blog posts presents new challenges in keeping track of the storyline or events that form the defining moments of the narrative. Several recent works \cite{dore2018theory,yuan2017machine,chung2017supervised,kovcisky2018narrativeqa,mostafazadeh2017lsdsem} have made efforts to advance the research in narrative comprehension. However, the development of computational models that automatically detect and interpret different structural elements of a narrative remains an open problem. Discovery of structural elements of a narrative has many applications in: (a) retrieval of narratives based on similar dramatic events or concepts instead of keywords \cite{mccabe1991developing,finlayson2006analogical,marchionini1991authoring}, (b) linking related stories that form a narrative thread towards theme generation \cite{berman2013relating}, (c) summarization of stories \cite{lehnert1981plot,papalampidi2020screenplay} and (d) story ending prediction or generation \cite{chen2019incorporating,li2013story,mostafazadeh2017lsdsem}, (e) commonsense reasoning \cite{goodwin2012utdhlt,gordon2011commonsense}, to list a few.

\begin{figure}[ht]
\includegraphics[width=\columnwidth]{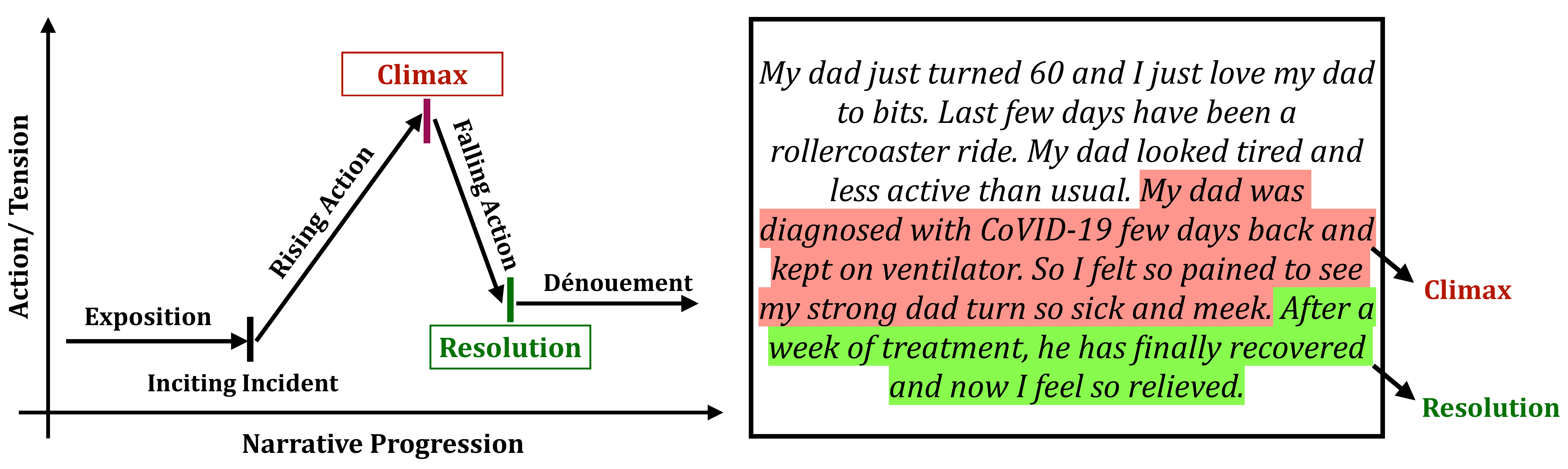}
\caption{\label{fig:prob_st} (Left) Freytag's Pyramid. (Right) Highlights of climax and resolution for a sample personal narrative.}
\vspace{-0.35cm}
\end{figure}

Several narrative theories have been proposed such as Freytag \cite{freytag1894technik}, Prince \cite{prince2012grammar}, Bruner \cite{bruner1991narrative,bruner2009actual}, Labov \& Waletzky \cite{labov1997narrative}, to name a few. These theories explain different elements of a narrative structure containing typical orderings between them. Certain elements of the narrative structure are correlated across different narrative theories. For example, Bruner's `breach in canonicity' \cite{bruner1991narrative} could correspond to (a) Freytag’s `climax' -- referring to the `turning point' of the fortunes of the protagonist \cite{abrams2014glossary} or (b) Labov's `most reportable event' (MRE) -- describing the event that has the greatest effect upon the goals, motivations and emotions of the characters (participants) in the narrative \cite{labov1997narrative,labov2006narrative}. Shorter narratives tend to consist mostly of complicating actions that culminate in the MRE or climax and instances of events that reach a `resolution' stage indicated by a swift drop in dramatic tension, while the other structural elements are more likely to occur in longer narratives. Figure \ref{fig:prob_st} (Left) shows Freytag's pyramid containing the key elements of the narrative structure and Figure \ref{fig:prob_st} (right) contains highlights of climax and resolution for a sample personal narrative. Thus, our work aims to leverage computational approaches at the intersection of information retrieval, NLP, and aspects of psychology, and model the key elements of narrative structure -- MRE and resolution. As an operating definition, we consider an MRE to be contained in a sentence(s) based on the following criteria -- it is an explicit event that can be reported as the summary of the story and occurs at the highest tension point of the story. Similarly, an event qualifies as `resolution' if it usually occurs after the MRE and resolves the dramatic tension in the narrative. 

Recently, Papalampidi et al. \cite{papalampidi2019movie} introduced a dataset consisting of movie screenplays and plot synopsis annotated with turning points. Few attempts have been made at annotating elements of high-level narrative structures \cite{li2017annotating} and automatically extracting them from free text. Ouyang et al. \cite{ouyang2015modeling}'s study on predicting MRE in narratives is the closest work to the problem considered in this paper. While most of these methods rely on syntactic, semantic, surface-level affect, or narrative features obtained using hand-engineering or pre-trained semantic embedding methods to model narrative structure, we investigate the role of a protagonist's psychological states in capturing the pivotal events in the narrative and their relative importance in identifying the elements of narrative structure -- Climax and Resolution. We find a basis for this study in prior theoretical frameworks \cite{murray2003narrative,ryan1986embedded,ouyang2014towards,lehnert1981plot,schafer2016exploring} that emphasize (a) how narrative structure organizes the use of psychological concepts (e.g. intentions, desires and emotions) and mediates all the human interactions and their social behavior, and (b) how protagonist's mental states (both implicit and explicit inferences, also imputed by readers) and psychological trajectory correlate with the classic dramatic arc of stories. Thus, to obtain the protagonist's mental states, we refer to a recent work \cite{vijayaraghavan2021modeling,sap2019atomic,rashkin2018modeling} that learns to embed characters' mental states using an external memory module. Our contributions are summarized below:

\begin{itemize}
    \item A $\textsc{Stories}$\footnote{Short for \textbf{ST}ructures \textbf{O}f \textbf{R}eddit P\textbf{E}sonal \textbf{S}tories} corpus containing a collection of Reddit Personal Narratives with fine-grained annotations of prominent structural elements of a narrative -- climax and resolution.
    
    \item An end-to-end neural network for modeling narrative structure, referred to as  $\textsc{M-sense}$\footnote{Short for \textbf{M}ental \textbf{S}tate \textbf{E}nriched \textbf{N}arrative \textbf{S}tructure mod\textbf{E}l}, that allows for integration of protagonist's mental state representations with linguistic information via multi-feature fusion.
    
    \item Experiments that analyze the impact of our modeling choices for short personal narratives. Specifically, we gauge the influence of incorporating mental state embeddings and report an improvement in $F_1$ scores of $\sim 11\%$ and $\sim 13\%$ over the base model for predicting climax and resolution respectively. 

\end{itemize}

\section{Related Work}

There is a large body of prior work that focuses on different aspects of narrative comprehension. Computational analysis of narratives operates at the level of characters and plot events. Examples include plot-related studies -- story plot generation, plot summarization, detecting complex plot units, modeling event schemas and narrative chains and movie question-answering; character-based studies -- inferring character personas or archetypes, analyzing inter-personal relationships and emotion trajectories, identifying enemies, allies, heroes;  story-level analysis -- story representation, predicting story endings, modeling story suspense, and creative or artistic storytelling, to list a few.

Several studies have analyzed the literature in narratology and formulated different goals and annotation labels associated with narratives for modeling their structure. 
Elson's \cite{elson2012detecting} Story Intention Graph (SIG) provided an annotation schema to capture timelines as well as beliefs, intentions, and plans of story characters. The annotations in this approach are similar to story generation methods described in Belief-Desire-Intention agents \cite{rao1995bdi} and intention-based story planning \cite{riedl2010narrative}. Previous studies like \cite{gordon2011commonsense} have analyzed personal weblog stories containing everyday situations. Rahimtoroghi et al. \cite{rahimtoroghi2014minimal} and Swanson et al. \cite{swanson2014identifying} used a subset of Labov's categories, including orientation, action, and evaluation in such personal weblog narratives. Black and Wilensky (1979) evaluate the functionality of story grammar in story understanding,
As explained earlier, \cite{papalampidi2019movie}'s dataset for analyzing turning points is a valuable addition in this area of work. Moreover, there have been consistent efforts \cite{jorge2018first,jorge20192} that study the link between Information Retrieval (IR) and narrative representations from a given text. These include works that exploit narrative structure in movies for IR \cite{jhala2008exploiting}, detect and retrieve narratives in health domain (patient communities \& medical reports) \cite{johnson2008electronic,rokach2004information,dirkson2019narrative,koopman2017generating}, identify narrative structures in news stories \cite{boyd2020narrative,levi2020compres} or generate summaries from screenplays or novels \cite{papalampidi2020screenplay}, to name a few. 
Given this wide spectrum of work, we leverage mental state representation models that are pre-trained using social commonsense knowledge aggregated using IR and text mining techniques. We employ the ensuing mental state embeddings in tandem with contextual semantic embeddings towards our primary objective of identifying elements of high-level narrative structure -- climax and resolution. We also conduct a detailed analysis of the outcome and the contribution of the protagonist's psychological state trajectory to our task.

\section{Dataset Collection}
\label{sec:dataset_coll}
\begin{figure}[ht]
\includegraphics[width=\columnwidth]{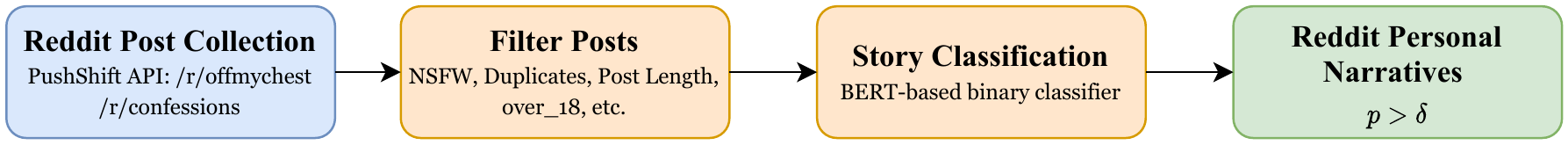}
\caption{\label{fig:dataset_coll} Illustration of our data collection pipeline.}
\vspace{-0.3cm}
\end{figure}

Figure \ref{fig:dataset_coll} presents our data collection pipeline. First, we collect Reddit posts from two communities: /r/offmychest and /r/confession using the PushShift API \footnote{https://pushshift.io/}. Next, we filter the collected data to retain only those posts that do not contain tags like ``[Deleted]'', ``NSFW'' \footnote{NSFW -- not safe for work} or ``over\_18''. Finally, we further narrow down the aggregated posts using a \textsc{Bert}-based story classifier. The pipeline is described in the Appendix.


\subsection{Annotation}
Here, we explain the annotation process involved in the construction of our $\textsc{Stories}$ dataset. Table \ref{tab:stories_stats} shows the descriptive statistics of our dataset.

\begin{table}[]
    \centering
    \small
    \begin{tabular}{p{0.15cm}r@{}}
\toprule
\multicolumn{2}{c}{\textbf{Dataset Statistics}}  \\ \midrule
\multicolumn{1}{l|}{\#Total Narratives}           & 63,258  \\
\multicolumn{1}{l|}{\#Annotated Narratives}           & 2,382  \\
\multicolumn{1}{l|}{\#Total Sentences}     & 42,614 \\
\multicolumn{1}{l|}{\#Climax Sentences}     & 5,173  \\
\multicolumn{1}{l|}{\#Resolution Sentences} & 4,502  \\ \bottomrule
\end{tabular}
    \caption{Statistics of our annotated $\textsc{Stories}$ dataset.}
    \label{tab:stories_stats}
\end{table}

\subsubsection{Setup}
We created a user interface for MTurk workers to make the annotation procedure convenient for capturing key elements of the narrative structure -- climax and resolution. The user interface allows the workers to highlight parts of the text that qualify as climax and resolution using red and green colors respectively. Three annotators were mainly involved in the annotation process. Each worker is presented a sampled text from the Reddit personal narrative corpus. Additionally, the workers are provided with an option of selecting checkboxes: ``No Climax'' or ``No Resolution''. This caters to those personal stories that don't contain a climax or resolution. 

\subsubsection{Agreements}
Once the data is collected using our annotation setup, we measure the inter-annotator agreement (IAA) at the sentence-level.  For sentence-level agreement, we use the following metrics: (i) Fleiss's kappa $(\kappa)$ \cite{fleiss1973equivalence}, (ii) mean annotation distance ($\mathcal{D}$), i.e., the distance between two annotations for each category, normalized by story length \cite{papalampidi2019movie}.

\subsubsection{Analysis}
\label{sec:dataset_analysis}
\begin{figure}[ht]
\includegraphics[width=\columnwidth]{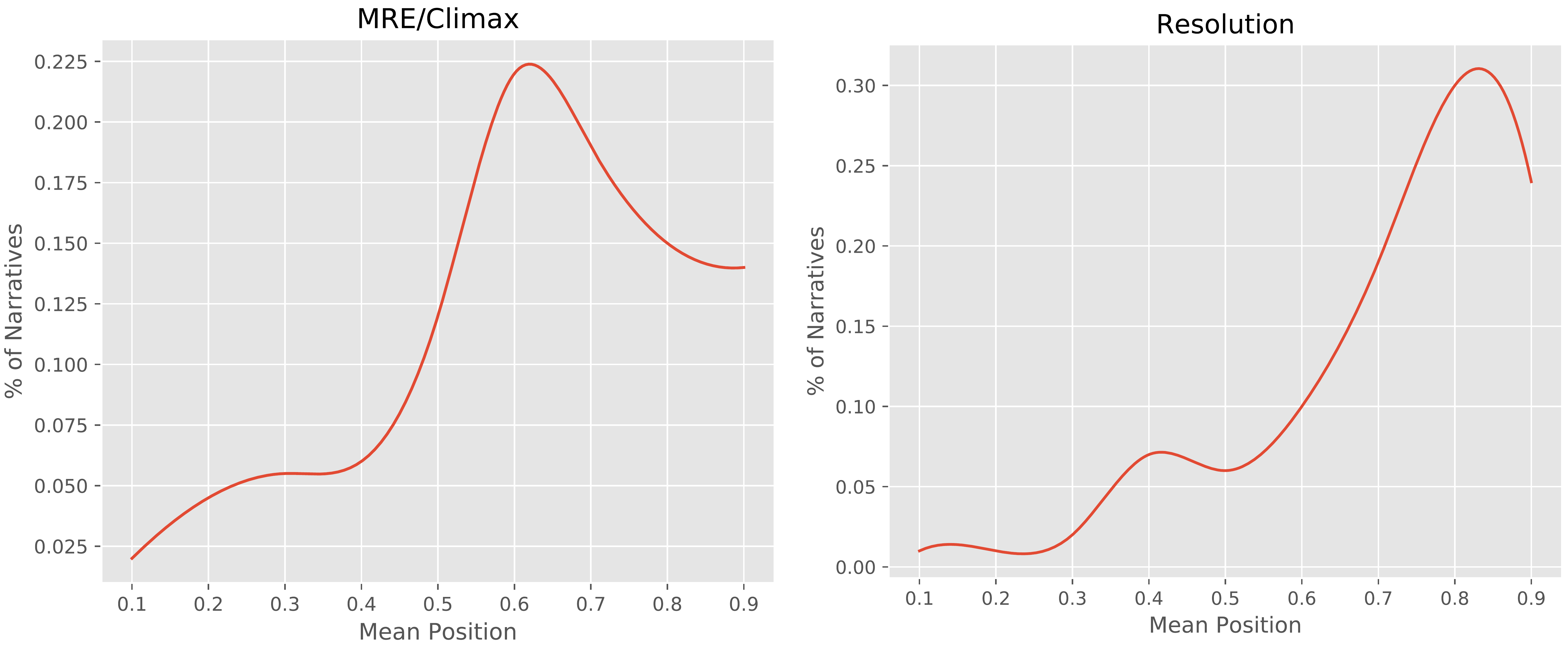}
\caption{\label{fig:data_distbn} Distributions of mean climax \& resolution sentence positions.}
\vspace{-0.45cm}
\end{figure}

\begin{figure*}[ht]
\includegraphics[width=\linewidth]{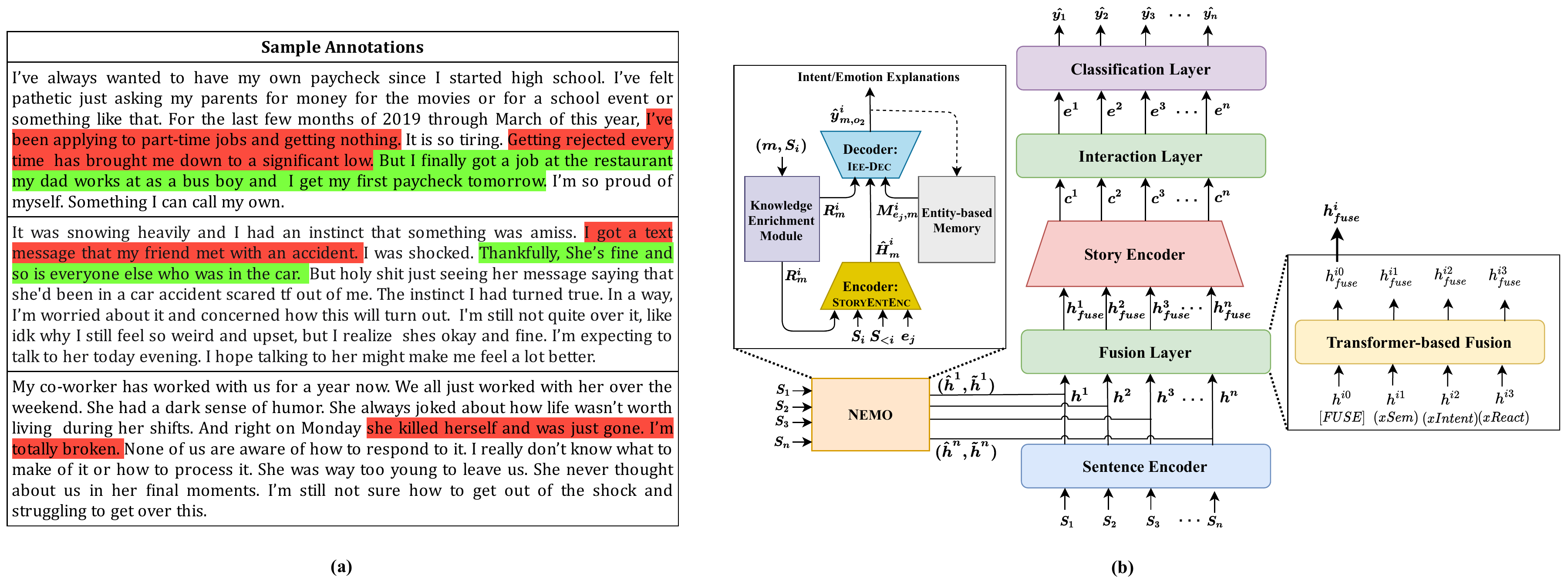}
\caption{\label{fig:model} (a) Sample annotations of climax (Red) and Resolution (Green) by one of the annotators. (b) Illustration of our $\textsc{M-sense}$ model. Note that $h^{i1}=h^i;h^{i2}=\hat{h}^i;h^{i3}=\tilde{h}^i$ relate to semantics $(xSem)$, intents $(xIntent)$ and emotional reactions $(xReact)$ of the $i^{th}$ sentence respectively. }
\vspace{-0.35cm}
\end{figure*}

We study the appearance of climax and resolution sentences by estimating their mean position normalized by the story length. We present the distribution of the position of both the structural elements in Figure \ref{fig:data_distbn}. While the average position for climax (0.61) coincides with the peak, we observe that the resolution contents occur later in the story. Table \ref{tab:iaa} shows the sentence-level IAA measures for each narrative element. We observe that substantial agreement is achieved for both the climax and resolution. Clearly, we obtain higher agreement values for resolution than the climax. Figure \ref{fig:model}a displays sample annotations (e.g. multi-sentence or non-contiguous highlights; no resolution) from our $\textsc{Stories}$ dataset. 


\begin{table}[]
\small
\centering
\begin{tabular}{@{}lcc@{}}
\toprule
\multicolumn{1}{c}{\textbf{Metric}}                                                   & \textbf{Climax} & \textbf{Resolution} \\ \midrule
Percentage Agreement         & 0.736     & 0.807     \\
Fleiss's Kappa $(\kappa)$            & 0.646     & 0.756     \\
\begin{tabular}[c]{@{}l@{}}Mean Annotation Distance $(\%   \mathcal{D})$\end{tabular} & 1.764           & 1.590               \\ \bottomrule
\end{tabular}
\caption{\label{tab:iaa} Sentence-level inter-annotator agreement.}
\vspace{-0.5cm}
\end{table}

\section{\textsc{M-sense}: Modeling Narrative Structure}
\label{sec:msense}
In this paper, we explore different modeling and analysis methods for understanding narratives and automatically extracting text segments that act as key elements of narrative structure, particularly climax and resolution. The models are provided a narrative text $T$ with $L$ sentences, $T=[S_1, S_2,..., S_L]$, as input. Here, each sentence $S_i$ contains $N_i$ words $\{w_1^i, w_2^i, .., w_{N_i}^i\}$ from vocabulary $\mathcal{V}$. Towards automatic detection of structural elements, we formulate it as a sentence labeling task where the goal is to predict a label $\hat{y}_i \in \{None, Climax, Resolution\}$ for each sentence $S_i$, based on the story context. 
Beyond linguistic features extracted from narratives, we focus on a dominant aspect in which a narrative is formed or presented, that is an account of characters' mental states -- motives and emotions. Thus, we leverage transfer learning from pretrained models trained to infer characters' mental states from a narrative. 
We implement a multi-feature fusion based learning model, $\textsc{M-sense}$, that potentially encapsulates syntactic, semantic, characters' mental state features towards our overall goal of predicting climax and resolution in short personal narratives. Our $\textsc{M-sense}$ model consists of the following components:

\noindent \textbf{Ensemble Sentence Encoders}, which computes per-sentence linguistic \& mental state embeddings.

\noindent \textbf{Fusion layer}, which integrates the protagonist's mental state information with the extracted linguistic features.

\noindent \textbf{Story Encoder}, which maps the fused encodings into a sequence of bidirectionally contextualized embeddings.

\noindent \textbf{Interaction layer}, which estimates state transition across sequential context windows to identify the boundaries. 

\noindent \textbf{Classification layer}, which involves linear layers to eventually calculate the label probabilities.

\subsection{Ensemble Sentence Encoders}
\label{sec:cne}

In this work, we aim to exploit both linguistic and mental state features for an enhanced model for narratives.
\subsubsection{Extracting Linguistic Representations} 
Pretrained general purpose sentence encoders usually capture a hierarchy of linguistic information such as low-level surface features, syntactic features and high-level semantic features. Given a narrative text with $L$ sentences $T=[S_1, S_2, ..., S_L]$, this component outputs hidden representations for sentences $H_{sents}=[h^1, h^2, ..., h^L]$ using different encoding methods.
In our $\textsc{Msense}$ model, we use a token-level $\textsc{Bert}$-based sentence encoder (more in the Appendix). Here, each sentence is prepended with a special $[CLS]$ token at the beginning of each sentence and appended with a $[SEP]$ token at the end of each sentence in the narrative. We apply both position and segment embeddings and feed to the pre-trained \textsc{Bert} model as: 
 \vspace{-0.15cm}
\begin{multline}
    \scalebox{0.85}{$H=[h_{[CLS]}^1,..,h_{N_1}^1,h_{[SEP]}^1,.., h_{[CLS]}^i, .., h_{N_i}^i,.., h_{[SEP]}^L]$}\\  \scalebox{0.85}{$= \textsc{Bert}(T)$}
\end{multline}
The hidden representation of the $i^{th}$ $[CLS]$ token from the top \textsc{Bert} layer is extracted as the semantic embedding of the $i^{th}$ sentence. However, we drop the subscript $[CLS]$ from $h^i_{[CLS]}$ and denote the output semantic embeddings as:  $H^{xSem}_{sents}=[h^1, h^2,..., h^L]$.

\subsubsection{Incorporating Protagonist's Mental Representation}
\label{sec:mental}

Prior studies have established how the progression of a story is as much a reflection of a sequence of a protagonist's motivation and emotional states as it is the workings of an abstract grammar \cite{palmer2002construction,mohammad2013once,alm2005emotional}. We follow a recent work \cite{vijayaraghavan2021modeling} that implements a \textsc{Nemo} model, a variant of a Transformer-based encoder-decoder architecture to embed and explain characters' (or entities’) mental states. We extract the embeddings of intents and emotional reactions of the protagonist for a given sentence in the narrative conditioning on the prior story context. Figure \ref{fig:model}b contains the overview of the $\textsc{Nemo}$ architecture. The computation of mental state embeddings are facilitated by a knowledge enrichment module that consolidates commonsense knowledge about social interactions and an external memory module that tracks entities' mental states. Using prior context $(S_{<i})$, entity $(e_j)$ and mental state attribute information ($m \in \{xIntent, xReact\}$ representing intent and emotional reaction respectively), we use the encoder, $\textsc{StoryEntEnc}(\cdot)$, in this trained model to obtain entity-aware mental state representation of the current sentence $S_{i}$. The encoding process in the \textsc{Nemo} model is given by:
\begin{multline}
    \scalebox{0.86}{$(\hat{H}_{xIntent}^{i},\tilde{H}_{xReact}^{i})=\textsc{StoryEntEnc}(S_{i}, S_{<i}, e_j, m);$} \\ 
    \scalebox{0.86}{$\forall m \in \{xIntent, xReact\}$}
    \vspace{-0.5cm}
\end{multline}
where $e_j \in \mathcal{E}$ is the entity, $(\hat{H}_{xIntent}^{i},\tilde{H}_{xReact}^{i})$ is the resulting entity-aware intent and emotion representation of the $i^{th}$-sentence given the story context. In this work, we use the narrator (``I'' or ``self'' in the personal narratives) as the protagonist. We only utilize the hidden representations of the $[CLS]$ token from both $(\hat{H}_{xIntent}^{i},\tilde{H}_{xReact}^{i})$ for subsequent processing steps. We denote these intent and emotion representation as: $H_{sents}^{xintent}=[\hat{h}^1,..,\hat{h}^L]$ and $H_{sents}^{xReact}=[\tilde{h}^1,..,\tilde{h}^L]$ respectively. 



\subsection{Transformer-based Fusion Layer}
 \label{sec:fusion}
Given multiple sentence-level embeddings, we apply a fusion strategy to derive a unified sentence embedding for our classification task. Let $h^{ik}; \forall{k \in \{1,...,K\}}$ denote different per-sentence latent vectors. In our case, $K=3$ and $h^{i1}=h^i;h^{i2}=\hat{h}^i;h^{i3}=\tilde{h}^i$ are embeddings related to semantics $(xSem)$, intents $(xIntent)$ and reactions $(xReact)$ of the $i^{th}$ sentence respectively. Drawing ideas from the literature of multimodal analysis \cite{urooj2020mmft}, we treat the multiple latent vectors as a sequence of features by first concatenating them together. We introduce a special token $[FUSE]$ \footnote{$[FUSE]$ is similar to the commonly used $[CLS]$ token.} that accumulates the latent features from different sentence encodings. The final hidden representation of $[FUSE]$ token obtained after feeding them to a Transformer layer is the fused output sentence representation: $h_{fuse}^i=\textsc{Tf}(\parallel_{k=0}^K{h^{ik}})$,
where \textsc{Tf} refers to the transformer encoder layer and $h^{i0}$ (i.e. when $k=0$) is set to the trainable $[FUSE]$ vector.

\subsection{Story Encoder}

We apply Transformer layers on the top of the sentence representations to extract narrative-level features. We refer it as \textit{Inter-sentence Transformer}. Intuitively, the transformer layer focuses on possibly different sentences in the narrative, and produces context-aware sentence embeddings. This is given as:
\begin{equation}
\begin{array}{c}
    \scalebox{0.8}{$\hat{H}^{l}=LayerNorm(\hat{C}^{l-1}+ \textsc{Mha}(\hat{C}^{l-1})$}\\
    \scalebox{0.8}{$\hat{C}^{l}=LayerNorm(\hat{H}^{l}+ \textsc{Ffl}(\hat{H}^l))$}\\
    \scalebox{0.8}{$C_{sents}=[c^1, c^2,..., c^L]=\hat{C}^{n_L}$}
\end{array}
\end{equation}
where $\hat{C}^0=PE(H_{sents})$, $PE$ refers to the positional encoding, $LayerNorm$ refers to layer normalization operation, \textsc{Mha} is the multi-head attention operation and \textsc{Ffl} is a feed-forward layer \cite{vaswani2017attention}. The superscript $l$ indicates the depth of the stacked Transformer layers. The output from the topmost layer, $l=n_L$, is our contextualized sentence embeddings $C_{sents}$. 

\subsection{Interaction layer}
\label{sec:interact}
In this layer, we compute the transition of state across sentences by measuring similarity metrics in the embedding space between sequential context windows and concatenating them with contextualized embeddings. By choosing windows of size $s$, we compute the left $(c_{left}^i)$ and right $(c_{right}^i)$ context information for the $i^{th}$ sentence by computing the mean sentence embedding within that window. Finally, we get the interaction-feature enhanced context-aware embeddings: $E_{sents}=[e^1, e^2,..., e^L]$. 

\subsection{Classification layer}
\label{sec:lin}
The resulting embeddings are mapped to a $C$-dimensional output using a softmax-based classification layer. Here, $C=3$ is the number of labels.
This step is given as: $\hat{y}_i=softmax(f_{s}(e^i))$.
 


\section{Experiments}
\label{sec:expt}
We conduct experiments to study the following research questions:

\textbf{RQ1:} How does our model compare with other baselines for identifying climax and resolution in short narratives?

\textbf{RQ2:} How do various model components contribute to the overall performance? To what extent do mental state representations play a role for our classification task?
\subsection{Overall Predictive Performance (RQ1)}

\subsubsection{Baselines}
We compare our model with a set of
carefully selected zero-shot (see the Appendix section) \& supervised baselines, shown as follows.

\textbf{Random baseline}, which assigns labels (Climax, Resolution or None) to sentences randomly.
    
    \textbf{Distribution baseline}, which picks sentences that lie on the peaks of the empirical distributions for climax and resolution in our training set as explained earlier.
    
    \textbf{Heuristic baseline}, which labels the sentences as climax or resolution based on heuristics. While we use the sentence that is the closest semantic neighbour of the post title as climax, the last sentence in the narrative is labeled as the resolution (as explained in the Appendix section.).
     
     \noindent A recent work \cite{wilmot2020suspense} has explored surprise a measure of suspense in narratives. \cite{ely2015suspense}'s surprise is defined as the amount of change from previous sentence to the current sentence in the narrative (see the Appendix). We encode the sentences in the story using the following approaches and eventually compute suspense measures for our classification task.  
    
    \noindent \textbf{GloVeSim} \cite{pennington2014glove}, \textbf{\textsc{Bert}} \cite{devlin2018bert}, \textbf{\textsc{Use}} \cite{cer2018universal}, which computes semantic embeddings using average word vectors (using GloVe) or Transformer-based models.
    
    \noindent \textbf{\textsc{StoryEnc}} \cite{papalampidi2019movie}, which uses the hierarchical RNN based language model to encode sentences in the story. 
    
    
    
    
   \noindent \textbf{\textsc{StoryEntEnc}} \cite{vijayaraghavan2021modeling}, which encodes the sentences in the story from the protagonist's perspective. Here, we denote intent and emotion embeddings as $(E_{int}=H_{sents}^{xIntent})$ and $(E_{emo}=H_{sents}^{xReact})$ respectively. 
    
   \noindent \textbf{\textsc{Cam}} and \textbf{\textsc{Tam}} \cite{papalampidi2019movie} consist of bidirectional LSTM model with the latter model having an additional interaction layer to compute boundaries between the topics in each story.
    
    
    \noindent \textbf{\textsc{M-sense--Fusion}}, which is a variant of our $\textsc{M-sense}$ model without mental state embeddings. 

\noindent \textbf{\textsc{M-sense}}, which is our complete model incorporating protagonist's mental representation.

    \begin{table}[]
\centering
\small
\begin{tabular}{@{}lcccc@{}}
\toprule
\multicolumn{1}{c}{\textbf{Models}}        & \multicolumn{2}{c}{$\boldsymbol{F_1\uparrow}$}               & \multicolumn{2}{c}{$\boldsymbol{D\downarrow}$} \\ \midrule
\multicolumn{1}{l|}{}                               & \textbf{C}     & \multicolumn{1}{c|}{\textbf{R}}     & \textbf{C}    & \textbf{R}    \\ \midrule
\multicolumn{1}{l|}{Random}                & 0.196          & \multicolumn{1}{c|}{0.143}          & 29.05         & 30.57         \\
\multicolumn{1}{l|}{Distribution} & 0.274          & \multicolumn{1}{c|}{0.315}          & \textbf{15.79}    & \textbf{14.42}   \\
\multicolumn{1}{l|}{Heuristic}             & 0.217          & \multicolumn{1}{c|}{0.147}          & 23.74         & 26.82         \\ \midrule
\multicolumn{1}{l|}{GloVeSim}                       & 0.312          & \multicolumn{1}{c|}{0.344}          & 12.06         & 11.65         \\
\multicolumn{1}{l|}{$\textsc{Bert}_{tok}$}    & 0.408          & \multicolumn{1}{c|}{0.441}          & 9.37          & 8.09          \\
\multicolumn{1}{l|}{$\textsc{Bert}_{sent}$} & 0.352          & \multicolumn{1}{c|}{0.366}          & 10.88         & 9.73          \\
\multicolumn{1}{l|}{$\textsc{Use}_{sent}$}  & 0.379          & \multicolumn{1}{c|}{0.391}          & 10.42         & 9.58          \\
\multicolumn{1}{l|}{$\textsc{StoryEnc}$}            & 0.410          & \multicolumn{1}{c|}{0.438}          & 8.81          & 7.46          \\
\multicolumn{1}{l|}{$E_{int}$}                      & \textbf{0.437} & \multicolumn{1}{c|}{0.462}          & \textbf{8.19} & 6.94          \\
\multicolumn{1}{l|}{$E_{emo}$}                      & 0.429          & \multicolumn{1}{c|}{\textbf{0.475}} & 8.43          & \textbf{6.67} \\ \midrule
\multicolumn{1}{l|}{$\textsc{Tam}$}                 & $0.565_{\pm0.022}$          & \multicolumn{1}{c|}{$0.609_{\pm0.008}$}          & 5.90$_{\pm3.18}$          & 5.02$_{\pm2.82}$          \\
\multicolumn{1}{l|}{$\textsc{Cam}$}                 & $0.578_{\pm0.019}$          & \multicolumn{1}{c|}{$0.604_{\pm0.0032}$}          & 6.58 $_{\pm4.02}$         & 5.44$_{\pm3.05}$          \\
\multicolumn{1}{l|}{$\textsc{M-sense}$}     & \textbf{0.694*}$_{\pm0.0027}$ & \multicolumn{1}{c|}{\textbf{0.743*}$_{\pm0.0015}$} & \textbf{4.15}$_{\pm1.84}$     & \textbf{3.20}$_{\pm1.06}$    \\ \bottomrule
\end{tabular}

\caption{\label{tab:main_res} Evaluation Results of different models to detect \textbf{C}limax and \textbf{R}esolution. We report $F_1$ score per class \& percent $(D)$ for these models. We use $\uparrow,\downarrow$ to indicate if higher/ lower values mean better performance respectively. * refers to significance $(p < 0.05)$ over TAM using a paired T-Test.}
\vspace{-0.5cm}
\end{table}

\subsubsection{Results}
Table \ref{tab:main_res} outlines the results of our evaluation. We report the performance of simple baselines, of which the distribution baseline turns out to be the strongest. The heuristic baseline performs slightly better than the random baseline. This suggests that the Reddit post title contains relevant signal to predict the climax while the last sentence heuristic for resolution is only as good as a random classifier. 

Applying suspense-based approaches with different sentence embedding methods yields relative improvement over the simple baselines in terms of both the evaluation metrics. As expected, sentence-level $\textsc{Bert}/\textsc{Use}$ performs worse than its token-level counterpart. We attribute this variation in performance to the lack of any story context information for computing latent embedding, thereby affecting the assessment of state changes in the narrative. However, sentence-level $\textsc{Use}$'s ability to produce better similarity estimates gives it a slight advantage over  sentence-level $\textsc{Bert}$. Notably, sentence representations obtained from models trained on stories $(\textsc{StoryEnc}, \textsc{StoryEntEnc})$ recorded comparable to improved results over other sentence embedding methods. Strikingly, computing surprise using protagonist mental state embeddings exhibit an overall enhanced classification capability. We find that the intent embedding $(E_{int})$ helps achieve the best zero-shot performance for detecting climax. A competitive outcome for resolution is obtained using protagonist's emotion representation $(E_{emo})$. We compare our complete $\textsc{M-sense}$ model with the best performing prior models such as $\textsc{Cam}, \textsc{Tam}$ \cite{papalampidi2019movie} applied for similar tasks. As we can see, supervised fine-tuning approaches easily beat the earlier results obtained using zero-shot methods. 
 Finally, our $\textsc{M-sense}$ model achieves an absolute improvement of $\sim20.07\%$ and $\sim22\%$ for climax and resolution prediction respectively. 

\begin{table}[]
\centering
\small

\begin{tabular}{@{}l|cc@{}}
\toprule
\multicolumn{1}{c|}{\textbf{Model Variants}} & \multicolumn{2}{c}{\textbf{$\boldsymbol{F_1} \uparrow$}} \\ \midrule
                                        & \textbf{C}           & \textbf{R}           \\ \midrule
\textbf{$\textsc{M-sense}$}              & \textbf{0.688}       & \textbf{0.738}       \\ \midrule
\textbf{Sentence Encoder Variants}      & \multicolumn{1}{l}{} & \multicolumn{1}{l}{} \\
\quad w/ Sentence-level $\textsc{Bert}$ & 0.665                & 0.709                \\
\quad w/ Sentence-level $\textsc{Use}$  & 0.677                & 0.726                \\ \midrule

\textbf{Story Encoder Variant}         &                      &                      \\
\quad w/o Story Encoder  & 0.620                & 0.653                \\
\quad w/ Inter-Sentence $\textsc{Rnn}$  & 0.659                & 0.705                \\ \midrule
\textbf{Interaction Layer Variant}     & \multicolumn{1}{l}{} & \multicolumn{1}{l}{} \\
\quad w/o Interaction Layer             & 0.654                & 0.716                \\ \midrule
\textbf{Fusion Layer Variants}          & \multicolumn{1}{l}{} &                      \\
\quad --w/o Fusion Layer                & 0.614                & 0.640                \\
\quad --w/o $E_{int}$                   & 0.638                & 0.703                \\
\quad --w/o $E_{emo}$                   & 0.652                & 0.687                \\ \bottomrule
\end{tabular}

\caption{\label{tab:ablation} We report $F_1$ score per class with non-default modeling choices for each component of our model.}
\vspace{-0.5cm}
\end{table}

\subsection{Ablation Study (RQ2)}
\label{sec:ablation}
To evaluate the contributions of each component in our $\textsc{M-sense}$ model, we conduct an ablation study using the validation set. For this study, we compare our best performing $\textsc{M-sense}$ model with alternative modeling choices for each of the components. Table \ref{tab:ablation} shows the results of our study. We modify one component at a time and report their performance using $F_1$ metric. This involves either replacing a component (denoted by ``w/'') or removing a component (denoted by ``w/o'' to refer without the component). For eg. ``w/ Sentence-level \textsc{Bert}'' refers to replacing token-level \textsc{Bert} in our $\textsc{M-sense}$ model with sentence-level \textsc{Bert} as our sentence encoder; ``w/o $E_{emo}$'' indicates the removal of protagonist's emotion state embedding from the fusion layer. 

\textit{Influence of Mental State Embeddings}: 
In this study, we examine the necessity of a fusion layer and probe the influence of protagonist's mental state embeddings towards our classification task. Notably, the results in Table \ref{tab:ablation} validate the benefits of introducing the fusion layer and demonstrate the relative performance gains obtained with intent and emotion embeddings. In the absence of a fusion layer, we observe that the performance drop is $\sim 11\%$ and $\sim 13\%$ for predicting climax and resolution respectively. The loss of the protagonist's intent information impacts the climax prediction more. This is analogous to the effect emotion information has on resolution prediction.

\subsection{Analysis and Discussion}
\label{sec:ana}

\begin{figure}[ht]
\centering
\includegraphics[width=0.7\columnwidth]{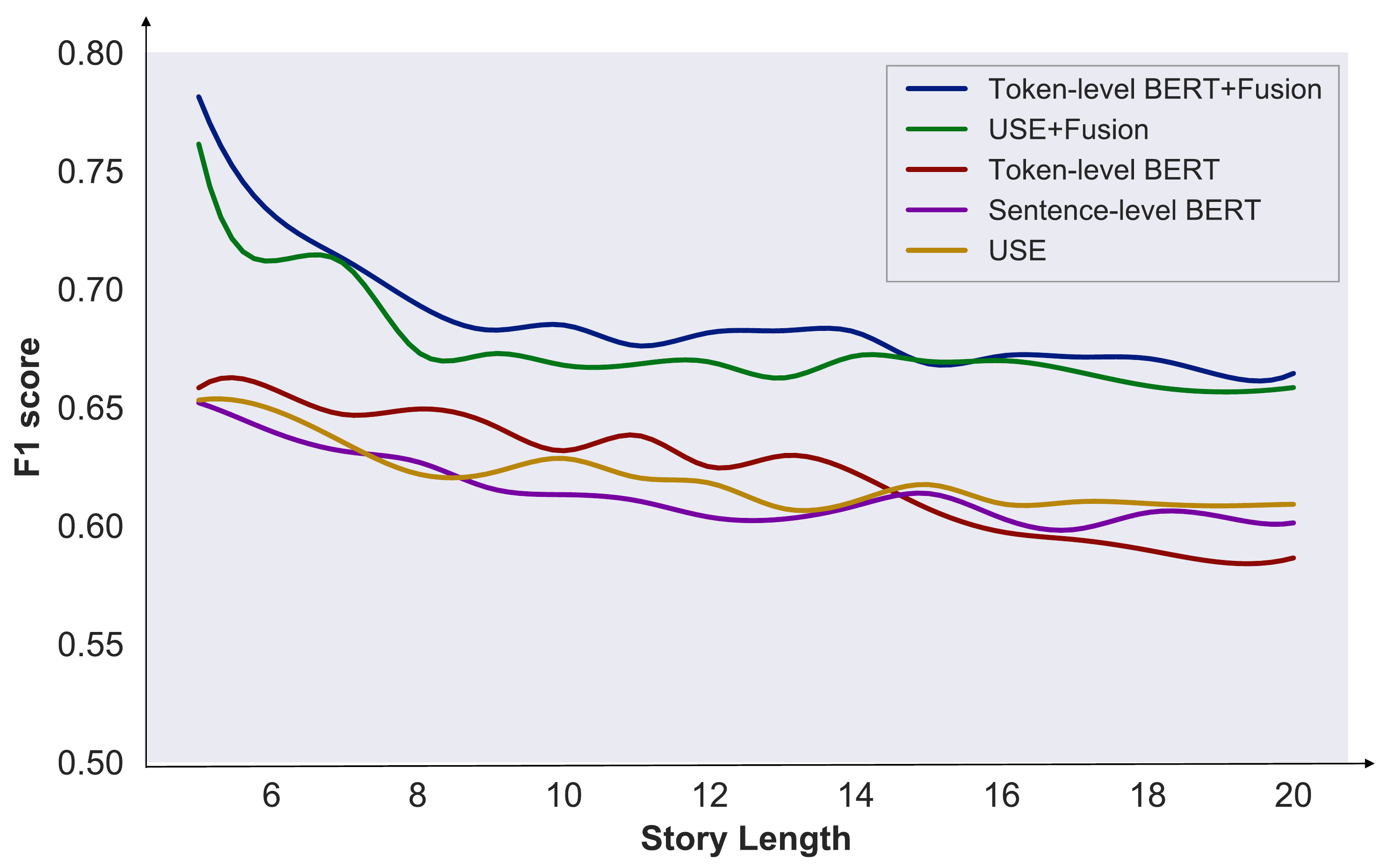}
\caption{\label{fig:sent_enc}  Performance of sentence encoders for detecting climax in story with varying length.}
\vspace{-0.5cm}
\end{figure}

\textit{Effect of Story Length}:
Here, we compare the performance of different sentence encoders with and without fusion layer for detecting climax in narratives with varying length. Figure \ref{fig:sent_enc} shows the results of this analysis. We observe that the token-level $\textsc{Bert}$ outperforms sentence-level $\textsc{Bert}$ and $\textsc{Use}$ encoders for narratives containing up to $13-14$ sentences, but the performance gradually degrades beyond 14 sentences. Sentence-level $\textsc{Use}$ encoder produces stable and relatively better outcomes for longer narratives (story length $> 14$). With the introduction of mental state representation through fusion layer, the $F_1$ score improved significantly irrespective of the sentence encoder used. 

\textit{Error Analysis}:
In order to estimate why our model augmented with mental state representation performs better, we conduct error analysis between our full \textsc{M-sense} model and the model without mental representation fusion ($\textsc{M-sense}-Fusion$). For those narratives where the latter model fails to predict correctly, we gauge the patterns emerging out of the following analysis: (a) Using VADER\footnote{https://github.com/cjhutto/vaderSentiment} \cite{hutto2014vader} a normalized, weighted composite sentiment score is computed for each sentence in the narrative, 
and (b) Using state classification \cite{rashkin2018modeling,vijayaraghavan2021modeling}, we assess Maslow's motivation or intent categories associated with sentences predicted as climax or resolution in the narrative and analyze for any pattern related to ground truth climax/resolution sentences. For predicting resolution, the $\textsc{M-sense}-Fusion$ makes $\sim28\%$ more mistakes than \textsc{M-sense} model for narratives with homogeneous endings (i.e. narratives having same sentiment sentences in the neighbourhood of resolution closer to the end of the story). $\textsc{M-sense}-Fusion$ model is unable to discern clearly and predicts a different sentence as resolution. Based on our analysis (b), there is a clear pattern that $\textsc{M-sense}$ gains significantly over the $\textsc{M-sense}-Fusion$ when the ground-truth climax sentences belong to ``Esteem'' and ``Love/Belonging'' categories. Our attention analysis results are shown in the Appendix section.

\section {Task: Modeling Movie Turning Points}

Given that our work is primarily focused on modeling narrative structure in personal narratives, we analyze how we can apply such a model towards identifying climax and resolution in movie plot synopsis. \cite{papalampidi2019movie} introduced a \textsc{Tripod} dataset containing a corpus of movie synopses annotated with turning points (TPs). By testing our model on this dataset, we evaluate our model's performance on an out-of-domain dataset. The dataset identified five major turning points in the movie synopses and screenplay, referring to them as critical events that prevent the narrative from drifting away. 
By their definitions for each of these categories \cite{papalampidi2019movie}, TP4 and TP5 align clearly with our usage of climax and resolution from prior narrative theories. Due to this alignment, it is relevant to use our model to predict these two categories in the \textsc{Tripod} dataset. However, we focus on the movie plot synopses in this work and use the cast information collected from IMDb as a part of this dataset.

We first apply our $\textsc{M-sense}$ trained on our $\textsc{Stories}$ corpus directly and evaluate its zero-shot performance (referred as $\textsc{Zs}$). We assume the protagonist in the movie to be the top character from the IMDb cast information. Though this may not always be true, it measures how our model fares on this dataset for predicting TP4 and TP5. Further, we use sentence-level \textsc{Use}-based sentence encoder as some of the wiki plot synopses are longer than what can be accommodated by our token-level \textsc{Bert} model. Additionally, we also fine-tune our model with the training set of the $\textsc{Tripod}$ dataset. This is denoted by $\textsc{M-sense}(FT)$.

\subsection{Results}
We display our model's performance compared to the best performing $\textsc{Tam}$ reported in the original work \cite{papalampidi2019movie}. $\textsc{Tam}$ with TP views implemented separate encoders for each of the categories and computed different representations for the same sentences acting as different views related to each TP. Similarly, $\textsc{Tam}$+Entities enriched the model with entity information by applying co-reference resolution and obtain entity-specific representations. Table \ref{tab:tripod} shows our model's results compared to the prior proposed approaches for modeling turning points in plot synopses. We find that our model in zero-shot settings outperforms a supervised $\textsc{TAM}$+TP views model, though it falls slightly behind the best supervised model.
Also, we restrict our model for predicting only two of the five major turning point labels. Finally, our fine-tuned model outperforms the best performing model, significantly reducing the mean annotation distance by an average of $\sim20\%$ on both the turning point labels. Thus, we are able to achieve remarkable improvement on an out-of-domain dataset even with assumptions on protagonist information. Therefore, we demonstrate that our $\textsc{Msense}$ model can predict climax and resolution in stories beyond just personal narratives, albeit limited by story length at this point.

\begin{table}[]
\centering
\small
\begin{tabular}{@{}lcc@{}}
\toprule
\textbf{Methods}       & \multicolumn{1}{l}{\textbf{TP4}}  & \multicolumn{1}{l}{\textbf{TP5}}  \\ \midrule
TAM+TP views           & 6.91                              & 4.26                              \\
TAM+Entities           & 5.23                              & 3.48                              \\
$\textsc{M-sense}(ZS)$  & 6.62                              & 4.54                              \\ \midrule
$\textsc{M-sense}(FT)$  & \multicolumn{1}{l}{\textbf{4.17}} & \multicolumn{1}{l}{\textbf{2.38}} \\ \bottomrule
\end{tabular}
\caption{\label{tab:tripod} Results of evaluation on $\textsc{Tripod}$ dataset. We report Mean Annotation Distance (\%) $D$ results for identifying TP4 and TP5 relevant to this work. }
\vspace{-0.5cm}
\end{table}

\section{Conclusion}

Towards modeling high-level narrative structure, we construct a dataset of personal narratives from Reddit containing annotations of climax and resolution sentences. 
Next, we introduce a deep neural model, referred to as $\textsc{M-sense}$, that learns to effectively integrate protagonist's psychological state features with linguistic information 
towards improved modeling of narrative structure. 
We experimentally confirm that our model outperforms several zero-shot and supervised baselines and benefits significantly from incorporating protagonist's mental state embeddings. Our model is able to achieve $\sim{20\%}$ higher success in prediction task than the previous methods.
We believe that our work will advance the research in understanding the larger dynamics of narrative communication and aid future efforts towards developing AI tools that can interact with users though stories. 
\bibliography{aaai23}
\clearpage

\appendix
\input{supplementary.tex}

\end{document}

%% file: supplementary.tex
\section{Appendix}
\subsection{Dataset Collection}

The first stage in this pipeline is dedicated to ingesting posts from Reddit. To collect natural first-person stories, we rely on Reddit communities comprising user-generated textual accounts of happy events, long-standing baggage, recent trauma, life experiences, adventurous encounters or guilt, and redemption episodes. To this end, we aggregate posts from two communities: /r/offmychest and /r/confession using the PushShift API \footnote{https://pushshift.io/}. The Pushshift API provides access to a database of all Reddit posts made since Reddit's launch as a social platform. We obtain $\sim440,000$ posts from this step.

Next, we filter the collected data to retain only those posts that do not contain tags like ``[Deleted]'', ``NSFW'' \footnote{NSFW -- not safe for work} or ``over\_18''. Relying on Prince's definition \cite{prince2012grammar} of a minimal story to comprise a starting state, a state-changing event, and an ending state, we eliminate posts containing less than three sentences. The subsequent stages in the pipeline are explained in the sections below.

\subsection{Story Classifier}

\label{sec:story_classifier}

The aggregated data consists of a wide variety of contents some of which do not qualify as personal narratives. In order to separate such non-narrative content from the collected data, we develop a story classifier that takes textual content as input and predicts the likelihood of the input text being a story. 

\subsubsection{Story vs. Non-Story Dataset} We gather a diverse collection of first-person blog text drawn randomly from the Spinn3r Blog Dataset containing everyday situations \cite{gordon2009identifying}.   Consistent with our filtering approach for Reddit posts, we follow a similar length criterion and sample $\sim 1,500$ blog posts. Further, we randomly selected $\sim 1,500$ texts from our Reddit posts corpus. Together, we obtain a total of $\sim 3,000$ posts to be annotated by MTurk workers.

For each post, annotators were instructed to read the textual content and choose one among the three labels: Story, Non-Story or Unsure \cite{gordon2013identifying}. We define these categories as follows -- (a) Story: Non-fictional narratives that people share with each other about their own life experiences. They contain a sequence of causally or temporally related events with the narrator being a participant; (b) Non-Story: Texts that are not primarily personal stories or don't give account of past events. They may or may not contain texts from first-person point of view but include opinion pieces, excerpts from news articles, recipes, technical explanations, facts, questions or some random discussion, personal advice, to list a few. When the annotators are uncertain about the right label, they are allowed to select the ``Unsure'' option. The three workers reached unanimous agreement on 76\% of the cases. We use the majority vote when such an agreement is not reached. Of the 3,000 posts, 1,197 posts were tagged as ``Story'', 1,173 as ``Non-Story'' and remaining as ``Unsure''.

\subsubsection{Model} 

We introduce a story classifier that separates non-fictional narratives from non-narrative textual content. Prior work have used feature engineering to extract features like Tf-Idf, Semantic Triplets, VerbNet \& coreference resolution chain based character features \cite{ceran2012semantic,eisenberg2017simpler} for this task. A work by Piper \cite{piper2018fictionality}  specifically used linguistic aspect of text to measure fictionality, i.e., distinguish works of fiction from non-fiction. In our work, we use a pretrained $\textsc{Bert}$ model for our classification task. Given an input text, the goal is to predict if the text qualifies as a story or not. We formulate the input text as $T = \{S_1, S_2, ..., S_n\}$, where $S_i$ is the $i^{th}$ sentence of the text. Following \cite{devlin2018bert}, we tokenize the input text and concatenate all tokens as a new sequence, $\{[CLS], S_1, [SEP], S_2, [SEP], ..., S_{n-1},$  $[SEP], S_{n},$ $[SEP]\}$, where $[CLS]$ is a special token used for classification and $[SEP]$ is a delimiter. Each token is initialized with a vector by summing the corresponding token and position embedding from pretrained \textsc{Bert}, and then encoded into a hidden state. Finally, we get $[H_{[CLS]}, H_{S_1},$ $H_{S_2}, .., H_{S_n},H_{[SEP]}]$ as an encoding output. We concatenate the $[CLS]$-token representation from the last four layers of the model for our classification task. We apply two linear layers on top of the concatenated output representation with a sigmoid activation function at the final linear layer. We optimize the binary-cross entropy loss and choose the model with least loss on the validation set as our final story classifier. We evaluate this model on the held-out test set. In Table \ref{tab:story_stats}a, we report the $F_1$ score, and compare our approach to other baselines. The best performing model achieves an $F_1$-score of 0.79. Finally, we feed the Reddit posts to the trained story classifier and obtain a probability score, $p$, that indicates the likelihood of the post being a story.
Furthermore, to increase the reliability of our data, we discard all those posts with probability score lesser that a chosen threshold $\delta$, i.e. $p<\delta$. In our work, we set $\delta$ to 0.75. This procedure yields a total of 63,258 stories, referred to as \textit{Reddit Personal Narratives} dataset.

%
\begin{table}[]
\centering
\small
\begin{tabular}{@{}lccc@{}}
\toprule
\multicolumn{1}{c}{\textbf{Models}} & \textbf{P} & \textbf{R} & \textbf{F1} \\ \midrule
Sem. Triplet \cite{ceran2012semantic}                & 0.64               & 0.47            & 0.46        \\
VerbNet \cite{eisenberg2017simpler}                 & 0.71               & 0.66            & 0.68        \\
$\textsc{Han}$ \cite{yang2016hierarchical}                                 & 0.75               & 0.73            & 0.74        \\ \midrule
$\textsc{Bert}$ \cite{devlin2018bert}                               & 0.81               & 0.78            & 0.79         \\ \bottomrule
\end{tabular}


\caption{\label{tab:story_stats} Performance of our BERT-based story classifier on the annotated dataset.}
\end{table}

\subsubsection{Agreement}
We analyze the discrepancies in the annotated data to gain insights about the potential challenges in the annotation process.
For climax, we note that the annotators get confused with sentences that involve events contributing to rising action (or Labov's complicating action) which eventually culminate in a climax (or Labov's MRE). Further, we observe that the differences are finer in many instances and hence harder to reliably detect. 
Though we achieve higher agreement on resolution category, the annotation gets less accurate with ambiguities in resolution and aftermath/endings, especially when narratives don't have a clear resolution. Interestingly, the annotators are able to discern between the two interest categories despite the high cognitive load and complexity involved in detecting them from unstructured user generated content. 

\subsection{M-SENSE Model}
\subsubsection{Token-BERT}
\label{sec:token_bert}
Token-level $\textsc{Bert}$: Since \textsc{Bert} produces output vectors that are grounded to tokens instead of sentences, we undertake an \textit{input processing step} that involves insertion of a special $[CLS]$ token at the beginning of each sentence and a $[SEP]$ token at the end of each sentence in the input sequence. we feed the entire narrative text $T$ to the input processing step to get the following sequence: $\{[CLS], w_1^1, w_2^1, .., w_{N_1}^1, [SEP],$ $[CLS], w_1^2, w_2^2, .., w_{N_2}^2, [SEP],.., [SEP],$ $[CLS], w_1^L, w_2^L,$ $ .., w_{N_L}^L, [SEP]\}$, where $w_j^i$ is the $j^{th}$ in $i^{th}$ sentence in the narrative. The multiple $[CLS]$ symbols will aggregate the features for sentences taking the context into consideration. Next, we apply alternating segment embeddings indicative of different sentences in our input textual narrative. Given a narrative with four sentences, $[S_1,S_2,S_3,S_4]$, we assign the segment embeddings as $[E_A, E_B, E_A,E_B]$.
\subsection{Zero-shot Approaches}
\label{sec:zs}
In addition to our $\textsc{M-sense}$ model and its variants, we experiment with zero-shot methods that utilize either simple heuristics or suspense-based approaches to model narrative structure. 

\subsubsection{Heuristic-based Approaches}

In addition to different modeling approaches, we experiment with simple heuristics for automatically labeling the sentences in story. This method assumes that the title of the Reddit post provides the summary of the post and hence, could refer to the MRE/climax of the narrative. We use a pretrained sentence embedding model and compute the semantic similarity between the each sentence in the narrative and the original title of the Reddit post. We compute $\textsc{Use}$ embeddings of sentences to identify the nearest neighbor of the post title and label it as the climax.
Next, we assign the last sentence as the resolution because it is more like for the cognitive tension to drop and reach a resolution at the near end of the narrative. 

\subsubsection{Suspense-based Approaches}
A recent work \cite{wilmot2020suspense} has explored surprise and uncertainty reduction as a measure of suspense in narratives  by considering sentences as the primary unit of processing. In our work, we mainly focus on surprise values based on consequential state change in narratives. Intuitively, a large change in any particular state indicates increased surprise at that point in the narrative. \cite{ely2015suspense}'s surprise is defined as the amount of change from previous sentence to the current sentence in the narrative. 
Peaks in such measures could reflect potential events where protagonist faces key obstacles and these may act as the defining moments of narrative structure. Thus, we examine different sentence embedding techniques to compute state changes including change in protagonist's mental state representations in our experiments to recognize suspenseful states. This will act as a relevant baseline to determine the effectiveness of semantic and mental state features.
  
\subsection{Training \& Hyperparameters}
\label{sec:training}
Though we approach narrative structure model as a sentence classification task, we divide the collected data based on number of narratives into train, validation and test sets at 70-10-20 split. In our sentence labeling task, each sentence will be accompanied with the entire narrative context. On the training set, we perform data augmentation by replacing sentences in narrative context with their paraphrases. The paraphrases are generated using a back-translation approach \cite{edunov2018understanding} based on pretrained English$\leftrightarrow$German translation model. We limit the number of such modified sentences to $20\%$ of the story length. We tune the hyperparameters using grid search and the best configuration is obtained based on validation set performance. Our best configuration consists of a two-layer story encoder $(n_L=2)$ and a single transformer-based fusion layer with 12 heads. For the window size in the interaction layer, it is set to two sentences, i.e., $s=2$. We also added a dropout with rate 0.2 in order to prevent overfitting. We optimize using Adam \cite{kingma2014adam} at a learning rate of $\alpha = 0.0001$ and a batch size of 32 with PyTorch \cite{paszke2019pytorch} being our model implementation framework. We report the score over an average of 3 runs. All our experiments were conducted on multiple NVIDIA Tesla P100 GPUs each with 12GB memory.

\subsection{Results}
We conducted ablation study in Section \ref{sec:ablation}. Table \ref{tab:ablation} shows the results of our study. We describe some of the aspects below.

\textit{Importance of Contextualization}: Next, we evaluate the contribution of a story encoder to our classification task. We observe that the performance drops significantly (by $\sim 10\%$ and $\sim 12\%$ for climax and resolution prediction respectively) without a story encoder. The importance of contextualizing the story sentences is established as we see a marked improvement of $\sim 8\%$ in the average overall performance with the introduction of an inter-sentence $\textsc{Rnn}$ story encoding layer. Still, this performance lags behind our default $\textsc{M-sense}$ setting with Transformer-based story encoding layer. We find the story encoder being relevant even if the inter-sentence dependencies are captured using the token-level $\textsc{Bert}$ model. We attribute this to the task-specific inter-sentence relationships being unearthed as we fine-tune our model.

\textit{Choice of Sentence Encoder}: From the results, it is clear that token-level \textsc{Bert} generally performs better the sentence-level \textsc{Bert} variant. This is unsurprising as the sentence-level approach produces embeddings without story context. Such an approach results in the loss of fine-grained inter-sentence token dependencies that the token-level $\textsc{Bert}$ can extract. Experiments also suggest that sentence-level \textsc{Use} model trained on textual similarity tasks results in better $F_1$ scores than the \textsc{Bert} counterpart. 

\textit{Impact of Interaction Layer}: The addition of an interaction layer yields an average $\sim 4\%$ gain in performance for identifying climax and resolution. The advantage of introducing an interaction layer has been studied in prior studies \cite{hearst1997text,papalampidi2019movie} and we find the performance improvement to be congruous with these studies.

\begin{figure}[ht]
\includegraphics[width=\columnwidth]{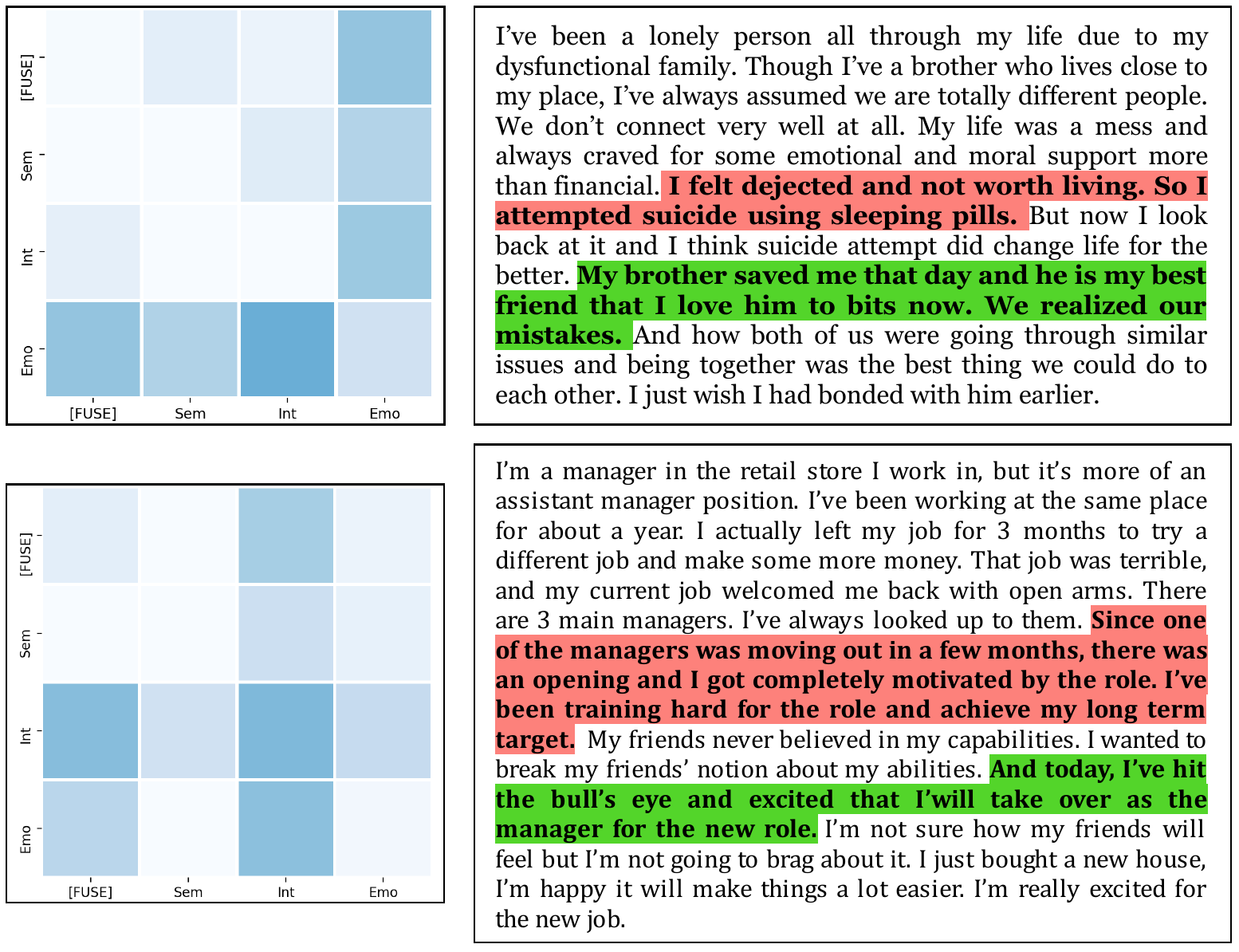}
\caption{\label{fig:attn_map} Attention analysis of two stories with climax and resolution sentences related to Maslow's categories -- Esteem (top) and Love/Belonging (bottom).}
\end{figure}

\subsection{Attention Analysis}
\label{sec:attn}
We conduct attention analysis on those narratives containing sentences belonging to ``Esteem'' and ``Love/ Belonging'' categories. Specifically, we study the functioning of Transformer-based Fusion layer for aggregating multiple latent embeddings -- Semantic $(Sem)$, Intent $(Int)$, and Emotion $(Emo)$. We then visualize the attention heatmaps from the fusion layer corresponding to these predicted climax and resolution sentences by computing the average attention score map over all heads. Figure \ref{fig:attn_map} displays the visualized attention map for sample stories belonging to the above mentioned categories. Since, $[FUSE]$ is the aggregated output over all the three latent embeddings. We note that the attention map has high attention scores between intent $(Int)$ and $[FUSE]$ vectors for stories related to ``Esteem'' motivation category, while more weight is assigned for emotion $(Emo)$ in samples associated with ``Love/Belonging'' category.